\documentclass[10pt, twocolumn, letterpaper]{article}

\usepackage{iccv}
\usepackage{times}
\usepackage{epsfig}
\usepackage{graphicx}
\usepackage{amsmath}
\usepackage{amssymb}
\usepackage{booktabs}
\usepackage{multirow}
\usepackage{lipsum}
\usepackage{comment}

\usepackage[pagebackref=true,breaklinks=true,letterpaper=true,colorlinks,bookmarks=false]{hyperref}

\iccvfinalcopy 

\def\iccvPaperID{6917} 

\ificcvfinal\fi

\newcommand{\Sum}[0]{\texttt{Sum}}
\newcommand{\Linear}[0]{\texttt{Linear}}
\newcommand{\MLP}[0]{\texttt{MLP}}
\newcommand{\GRU}[0]{\texttt{GRU}}

\newcommand\blfootnote[1]{%
  \begingroup
  \renewcommand\thefootnote{}\footnote{#1}%
  \addtocounter{footnote}{-1}%
  \endgroup
}

\begin{document}

\title{Long Short View Feature Decomposition via Contrastive Video Representation Learning}

\author{Nadine Behrmann$^1$ \qquad Mohsen Fayyaz$^{2,\dagger}$ \qquad Juergen Gall$^2$ \qquad Mehdi Noroozi$^1$\\
$^1$Bosch Center for Artificial Intelligence \qquad $^2$University of Bonn\\
{\tt\small \{nadine.behrmann, mehdi.noroozi\}@de.bosch.com, \{fayyaz, gall\}@iai.uni-bonn.de}
}

\maketitle
\ificcvfinal\thispagestyle{empty}\fi

\blfootnote{$^\dagger$ Work done during an internship at the Bosch Center for Artificial Intelligence.}

\begin{abstract}
    Self-supervised video representation methods typically focus on the representation of temporal attributes in videos.
    However, the role of stationary versus non-stationary attributes is less explored: Stationary features, which remain similar throughout the video, enable the prediction of video-level action classes. Non-stationary features, which represent temporally varying attributes, are more beneficial for downstream tasks involving more fine-grained temporal understanding, such as action segmentation.
    We argue that a single representation to capture both types of features is sub-optimal, and propose to decompose the representation space into stationary and non-stationary features via contrastive learning from long and short views, \ie long video sequences and their shorter sub-sequences. 
    Stationary features are shared between the short and long views, while non-stationary features aggregate the short views to match the corresponding long view.
    To empirically verify our approach, we demonstrate that our stationary features work particularly well on an action recognition downstream task, while our non-stationary features perform better on action segmentation.
    Furthermore, we analyse the learned representations and find that stationary features capture more temporally stable, static attributes, while non-stationary features encompass more temporally varying ones.
\end{abstract}

\section{Introduction}
\begin{figure}
\begin{center}
\includegraphics[width=\linewidth]{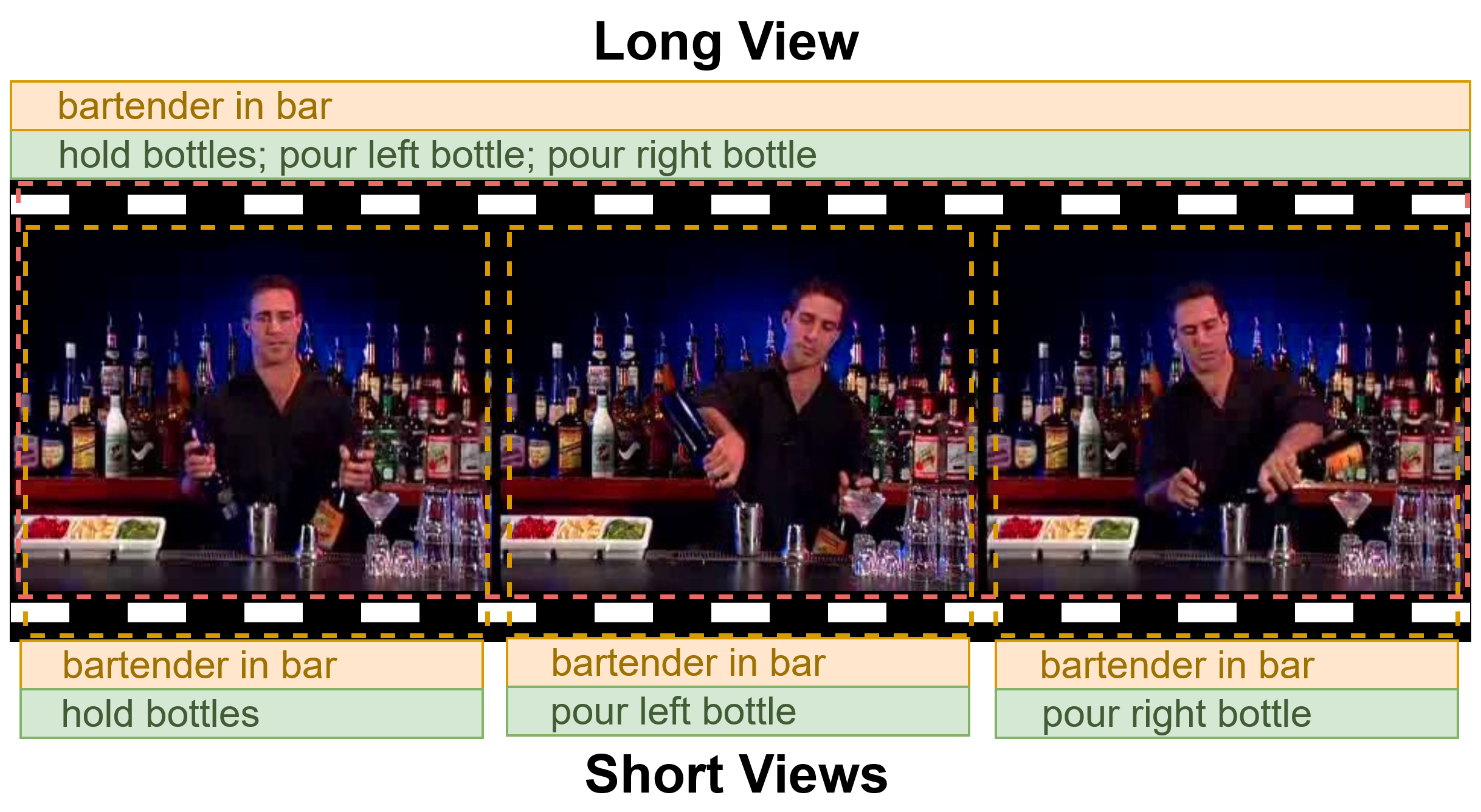}
\end{center}
\vspace{-0.5em}
\caption{Video attributes can be divided into two sections: Stationary features (shown in yellow), that are shared between long and short views, and non-stationary features (shown in green), that aggregate the short views to match the corresponding long view.}
\label{fig:teaser}
\vspace{-0.5em}
\end{figure}

Learning rich video representations is a key challenge for general video understanding. An ideal representation extracts useful information that benefits numerous downstream tasks such as action recognition, retrieval and action segmentation.
Learning such representations in a supervised setting is inherently biased towards static features~\cite{Li:2018:RESOUND}. 
However, in order to solve more complex downstream tasks, that require temporal attributes of videos such as temporal action segmentation, we need a more diverse set of features.
As a remedy, we train our network to represent stationary and non-stationary features.
To get an intuition, let's consider the following example:  a video of a bartender mixing a cocktail in Figure~\ref{fig:teaser} is stationary in one sense -- we see a bartender in a bar with liquor bottles in the background -- but non-stationary in another: different steps in preparing a cocktail are shown, such as holding the bottle, pouring the left bottle and pouring the right bottle.
Here, the stationary attributes of the video enable us to predict the overall action class, \ie mixing cocktails. On the other hand, the non-stationary attributes enable more fine-grained temporal distinctions, \eg predicting when different steps occur in the video.
Ideally, both types of attributes should be represented by video models.

Learning representations in a supervised setting, usually involves pretraining on large-scale labeled datasets such as Kinetics~\cite{Kay:2017:Kinetics} with video-level annotations, inhibiting strong static biases~\cite{Li:2018:RESOUND}. As a result they capture mainly the stationary features, but largely ignore non-stationary features as the stationary features are sufficient for action recognition.
Self-supervised leaning provides a promising direction to address this shortcoming.
As the supervisory signals arise from the underlying structure of the data, it has the potential to extract more descriptive features.
Several previous self-supervised methods aim to capture temporal features in videos by designing a temporal pretext task, \eg predicting temporal transformations~\cite{Jenni:2020:VRL} or video speed~\cite{Benaim:2020:SpeedNet}. 
These methods are not explicitly encouraged to capture stationary and non-stationary features.
In contrast, we explicitly decompose the representation space into stationary and non-stationary features, enabling us to solve a more diverse set of downstream tasks, including action recognition and temporal action segmentation.

Following the recent trend in self-supervised learning, our proposed method fits in the contrastive learning framework. The supervisory signal that leads us to distinguish between the stationary and non-stationary features emerges from long and short views of a given video. Naively applying contrastive learning to long and short views results in a set of features that represent both long and short views similarly. We argue that this assumption is only valid for a subset of features, which we call stationary. The other subset, which we call non-stationary features, includes a set of features that aggregate from short to long views, \ie combining attributes of several short views gives us the attributes of the long view, see Figure~\ref{fig:teaser}. Therefore, imposing a naive similarity between the long and short views is prone to ignoring the non-stationary features, which are crucial for different downstream tasks. Accordingly, we divide the final features into two disjoint subsets, which are later used in two separate contrastive losses. For the stationary features of a given long view,  we provide a positive pair through the stationary features of a corresponding short view. Whereas an aggregated form of the non-stationary features of all corresponding short views forms a positive pair for the non-stationary features of the long view.

We validate our argument above empirically and train our method on the Kinetics dataset~\cite{Kay:2017:Kinetics} without using any labels. We evaluate the model on several downstream tasks and datasets and analyse the learned representations; we highlight the main results here.
Our proposed method achieves state-of-the-art retrieval results on the HMDB51 dataset, and outperforms a contrastive learning baseline using data augmentation by significant margins on the UCF101~\cite{Soomro:2012:UCF} and HMDB51~\cite{Kuehne:2011:HMDB} datasets for action classification: $78.0\%$ and $53.7\%$ accuracy versus $72.7\%$ and $46.3\%$, respectively.
Additionally, we evaluate the learned representations for temporal action segmentation on the Breakfast dataset~\cite{Kuehne:2014:Breakfast}. 
We show that, our non-stationary features perform substantially better than the stationary features on this dataset, supporting our hypothesis on the design choice. To our knowledge, we are the first to conduct this evaluation for video representation learning.
Furthermore, we analyse the learned representations and find that stationary features capture attributes, that remain similar over time and can be detected in a few frames, while non-stationary features encompass more temporally varying attributes, which are revealed when observing more frames.

Our contributions are as follows: 
1) We propose the novel approach of \textbf{L}ong \textbf{S}hort View \textbf{F}eature \textbf{D}ecomposition (LSFD).
2) Our proposed method achieves state-of-the-art retrieval performance on HMDB51 using a 3D-Resnet18. 
3) We evaluate our proposed method on a temporal action segmentation task for the first time, and show that it outperforms the supervised baseline with high margins across multiple metrics. 
4) We investigate the feature decomposition capability of our method and analyse the learned representations. We find that stationary features remain more stable over time and perform better on the tasks and datasets that include large static biases, \eg action classification on UCF101. On the other hand, non-stationary features vary over time and are beneficial for more dynamic tasks and datasets, \eg temporal action segmentation on the Breakfast dataset.

\section{Related Work}
\paragraph{Pretext tasks for videos.}
While image-level pretext tasks can be extended to videos \cite{Jing:2018:SSS, Kim:2019:STC}, the temporal structure in videos provides further opportunities for this purpose.
For example, a supervision signal can be derived from the order of frames~\cite{Misra:2016:SAL} or video clips~\cite{Xu:2019:VCOP}.
Temporal context can be used to create a cloze procedure~\cite{Luo:2020:VCP}.
Other tasks are based on the frame rate, such as speed prediction~\cite{Benaim:2020:SpeedNet}, pace prediction~\cite{Wang:2020:PP}, playback rate perception~\cite{Yao:2020:PRP}, or predicting temporal transformations~\cite{Jenni:2020:VRL}.

\paragraph{Contrastive learning.}
Originally proposed in~\cite{Dosovitskiy:2016:DUF}, instance recognition has become the underlying principle for many modern contrastive methods.
The contrastive loss, which was first proposed in~\cite{Sohn:2016:Npair} and later popularized as \textit{InfoNCE} loss by~\cite{Oord:2018:CPC}, involves positive and negative pairs of features. It aims to maximize the similarity of positive pairs -- which are obtained by generating different views of the same data -- and minimize the similarity of negative pairs.
Views can be obtained from different channels~\cite{Tian:2019:CMC}, counting~\cite{Counting}, permutation~\cite{Misra:2019:PIRL}, or via augmentations.

The augmentations used to construct views, such as those explored in~\cite{Chen:2020:SimCLR}, have a substantial impact on the learned representation. For example, heavy cropping leads to occlusion-invariant representations~\cite{Purushwalkam:2020:DCS}.
To prevent shortcuts via low-level image statistics (\eg edges, corners, etc.), a challenging set of negatives is required. This is available via a memory bank~\cite{He:2019:MoCo}, which enables the storage of a large set of of negative samples, or hard negative mining.
For example, \cite{Han:2019:DPC} and the follow-up work \cite{Han:2020:MemDPC} obtain hard negatives from different spatio-temporal locations in the feature map, while \cite{BidirectionalFeaturePrediction} constructs negatives based on past and future clips.
More recently, \cite{Chen:2020:SimSiam} and \cite{Grill:2020:BYOL} have demonstrated that negatives can be omitted entirely.
The role of hard positives is investigated in~\cite{Han:2020:CoCLR}. 
Our method can be interpreted similarly: By pairing long views with short or aggregated views, the difficulty of the task lies within the positive pairs, which require the model to make a connection in terms of stationary and non-stationary features.

\begin{figure*}[ht]
\begin{center}
\includegraphics[width=0.9\linewidth]{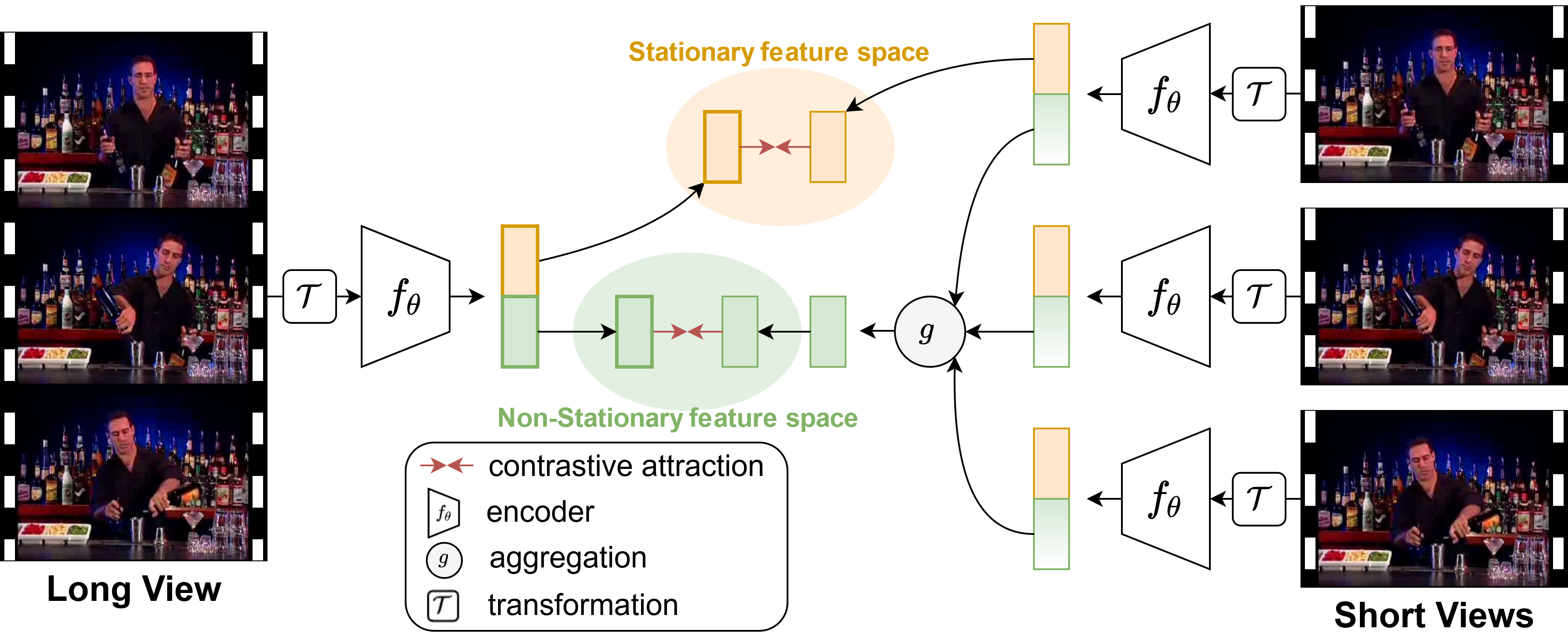}
\end{center}
   \caption{
   We extract features from long and short views and decompose them into stationary and non-stationary features, shown in orange and green, respectively. Stationary features remain similar over time and are shared by the long and short view, and serve as a positive pair (indicated by red arrows). Non-stationary features on the other hand capture temporal variances; we aggregate non-stationary features of short views to obtain the positive for the long view.}
\label{fig:MethodOverview}
\end{figure*}
\section{Method}

The key ideas of our \textbf{L}ong \textbf{S}hort View \textbf{F}eature \textbf{D}ecomposition (LSFD) method lie in decomposing the video representation into stationary and non-stationary features, and constructing short and long views of the data, see Figure~\ref{fig:MethodOverview}.  Short views provide local attributes, as they span a limited temporal receptive field, while global temporal attributes are better perceived through the larger temporal receptive field of long views. 
Therefore, a naive solution of imposing similarity between short and long views is not necessarily optimal.
We propose to establish a connection between long and short views by decomposing the representation space into two sections: 
One section represents stationary features that are shared between short and long views. The other section represents the attributes that should aggregate short views to the corresponding long view, which we call non-stationary features. As a result, the network is allowed to establish a proper connection between short and long views, without being forced to represent them similarly. We impose the concept of stationary and non-stationary features via separate contrastive losses on each section of the feature space. In the following, we discuss different components and design choices of our method in more detail.

\subsection{Stationary and Non-Stationary Features}
For a given sequence of video frames, we obtain a \textit{long} view $x_l$ that consists of all frames, and $N$ non-overlapping sub-sequences, $x_s^{(1)},\dots, x_s^{(N)}$, which serve as \textit{short} views. 
This allows us to construct positive pairs for stationary and non-stationary features, see Figure~\ref{fig:MethodOverview}.

More precisely, we train a parametric function $f_\theta$ that takes a sequence of videos frames and maps them to a representation space: $f_\theta(x) = \xi = (\psi, \phi)$, where $\psi,\phi$ denote the stationary and non-stationary features of $x$, respectively. 
We compute the features of the long and short views by feeding them to a shared backbone: 
\begin{align*}
    f_\theta(x_l) &= \xi_l = (\psi_l, \phi_l), \\
    f_\theta(x_s^{(i)}) &= \xi_s^{(i)} = (\psi_s^{(i)}, \phi_s^{(i)}) \quad \text{for } i\in\{1,\dots,N\}.
\end{align*}

This allows us to establish a connection between long and short views.
Stationary features represent attributes that remain the same over time, while non-stationary features are aggregated over time.
Overall, they satisfy the following two properties:
\begin{align}
\label{sim:static}
    \psi_l &\simeq \psi_s^{(i)} \quad  \text{for } i\in\{1,\dots,N\},  \\
\label{sim:dynamic}
    \quad \phi_l &\simeq g(\phi_s^{(1)},\dots, \phi_s^{(N)}).
\end{align}
Here, the aggregation function $g$ can be any function that takes the non-stationary features of the short views and maps them to non-stationary features of the long view.
Choosing an appropriate aggregation function is non-trivial and deserves extensive investigation.
Candidates range from simple functions, such as a sum, to more elaborate and learnable functions, for example a linear transformation, MLP, or a recurrent network. 
We provide an ablation of different aggregation functions in Section~\ref{sec:aggregation_ablation}.

In order to enforce the similarities in Eqs.~\eqref{sim:static} and \eqref{sim:dynamic} during training, we construct the following positive pairs to be used in two separate contrastive losses. 
The first pair $(\psi_s^{(j)}, \psi_l)$ aims at features shared between the long and a short view according to Eq.~\eqref{sim:static}. The second pair $ (g(\phi_s^{(1)},\dots, \phi_s^{(N)}), \phi_l)$ corresponds to Eq.~\eqref{sim:dynamic}, which aggregates the non-stationary features of the short views to match the non-stationary features of the long view via an aggregation function $g$.

For any given pair of features $(z_1,z_2)$ obtained through the described procedure above, we compute the similarity following the recent trend in contrastive learning \cite{Chen:2020:SimCLR}: We apply a learnable transformation $h$, here an MLP head, and a temperature parameter $\tau$ to scale the cosine similarity, denoted by
\begin{align}
    \text{sim}_{h}(z_1, z_2) = \frac{1}{\tau} \frac{h(z_1)^T h(z_2)}{\|h(z_1)\| \|h(z_2)\|}.
\end{align}

\subsection{Training Objective}
\label{sec:training_objective}
Our training objective consists of three separate InfoNCE losses applied on stationary, non-stationary, and full features: 
\begin{equation}
    \mathcal{L} = \mathcal{L}_{\text{stationary}} + \mathcal{L}_{\text{non-stationary}} + \mathcal{L}_{\text{instance}},
\end{equation}
which we will discuss below. We use three different MLP heas $h_s$, $h_n$, and $h_i$ for the three separate losses.
We use a set of negatives that consists of random videos:
\begin{align*}
    \mathcal{N} = \{f_\theta(\bar x_l) = \xi_\text{neg} = (\psi_{\text{neg}}, \phi_{\text{neg}})|\bar x_l \text{ is a random video}\}.
\end{align*}
To avoid shortcuts via low-level video statistics, we apply the same set of standard video augmentations, including random resized crop, horizontal flip, and color augmentation, to long and short views independently.

\paragraph{Stationary Loss.} Attributes that remain the same over time -- such as non-moving objects or the background scene
-- are shared between the long view and all short views; and thus should be represented similarly.
Therefore, the stationary features of a short view should capture the same attributes as the stationary features of the long view. We encourage such a property by applying the following loss function:

\begin{align}
   \label{eq:static}
    \mathcal{L}_{\text{stationary}} = 
    - \log \frac{\exp(\text{sim}_{h_{s}}(\psi_s^{(j)}, \psi_{l}))}{\displaystyle\sum_{\bar\psi_{l}\in \mathcal{N_\psi}\cup\{\psi_l\}}\exp(\text{sim}_{h_s}(\psi_s^{(j)}, \bar\psi_{l}))}.
\end{align}
Here, $\psi_s^{(j)}$ are the stationary features of a randomly selected short view, and $\mathcal{N}_\psi=\{\psi_{\text{neg}}|(\psi_{\text{neg}}, \phi_{\text{neg}})\in \mathcal{N} \}$.
  
\paragraph{Non-Stationary Loss.}
Complementary to stationary features, the non-stationary features represent the content of the video that varies: moving objects and persons, motion, temporal changes in the scene, etc.
These temporal changes \textit{aggregate} over time, \ie the non-stationary features of the long view should capture the temporal changes happening in all sub-sequences. We encourage such a property by applying the following loss function:

\begin{align}
   \label{eq:dynamic}
    \mathcal{L}_{\text{non-stationary}} = 
    - \log \frac{\exp(\text{sim}_{h_n}(\phi_g, \phi_{l}))}{\displaystyle\sum_{\bar\phi_{l}\in \mathcal{N_\phi}\cup \{\phi_l\}}\exp(\text{sim}_{h_n}(\phi_g, \bar\phi_{l}))}
\end{align}
where $\phi_g = g(\phi_s^{(1)},\dots, \phi_s^{(N)})$ is an aggregated version of the short view non-stationary features, and $\mathcal{N}_\phi~=~\{\phi_{\text{neg}}|(\psi_{\text{neg}}, \phi_{\text{neg}})\in \mathcal{N} \}$.

\paragraph{Instance Recognition Loss.} 
Finally, we add an instance recognition loss on the long views by applying InfoNCE on the full features.
To that end we obtain a second view of the long video sequence $\hat\xi_l$ via standard video augmentations. This corresponds to the standard contrastive learning approach.

\begin{align}
   \label{eq:vanilla}
    \mathcal{L}_{\text{instance}} = 
    - \log \frac{\exp(\text{sim}_{h_i}(\xi_l, \hat\xi_l))}{\displaystyle\sum_{\bar\xi_{l}\in \mathcal{N}\cup\{\hat\xi_l\}}\exp(\text{sim}_{h_i}(\xi_l, \bar\xi_{l}))}.
\end{align}

\section{Experiments}
\begin{table}
\begin{center}
    \begin{tabular}{ll|cc}
    \toprule
    \multicolumn{2}{c|}{Self-Supervised Methods}&\multicolumn{2}{c}{top1 Accuracy}\\
    Method & Arch & UCF & HMDB\\
    \midrule
    $\text{Shuffle\&Learn}^\dagger$ \cite{Misra:2016:SAL} & CaffeNet & 50.2 & 18.1 \\
    $\text{OPN}^{\dagger}$ \cite{Lee:2017:URL} & VGG & $59.8$ & $23.8$ \\
    $\text{VCOP}^{\dagger}$ \cite{Xu:2019:VCOP} & R3D & $64.9$ & $29.5$\\
    $\text{PRP}^{\dagger}$ \cite{Yao:2020:PRP} & R3D & $66.5$ & $29.7$ \\
    BFP \cite{BidirectionalFeaturePrediction} & 2D3D-R18 & $66.4$ & $45.3$ \\
    DPC \cite{Han:2019:DPC} & 2D3D-R34 & $75.7$ & $35.7$ \\
    MemDPC \cite{Han:2020:MemDPC} & 2D3D-R34 & $78.1$ & $41.2$ \\
    $\text{CBT}^{\ddagger}$ \cite{Sun:2019:CBT} & S3D & $79.5$ & $44.6$ \\
    SpeedNet \cite{Benaim:2020:SpeedNet} & S3D-G & $81.1$ & $48.8$ \\
    CoCLR \cite{Han:2020:CoCLR} & S3D-G & $87.9$ & $54.6$ \\ 
    \midrule
    3D-ST-Puzzle \cite{Kim:2019:STC} & C3D & $61.2$ & $28.3$ \\
    MA Stats \cite{Wang:2019:MA} & C3D & $61.2$ & $33.4$ \\
    $\text{Temp Trans}^{\ddagger}$ \cite{Jenni:2020:VRL} & C3D & $69.9$ & $39.6$ \\
    LSFD (Ours) & C3D & $\mathbf{79.8}$ & $\mathbf{52.1}$ \\
    \midrule
    $\text{3DRot}^{\ddagger}$ \cite{Jing:2018:SSS} & 3D-R18 & $62.9$ & $33.7$\\
    3D-ST-Puzzle \cite{Kim:2019:STC} & 3D-R18 & $65.8$ & $33.7$\\
    VIE \cite{Zhuang:2020:VIE} & 3D-R18 & $72.3$ & $44.8$ \\
    LA-IDT \cite{Tokmakov:2020:LA_IDT} & 3D-R18 & $72.8$ & $44.0$ \\
    $\text{Temp Trans}^{\ddagger}$ \cite{Jenni:2020:VRL} & 3D-R18 & $\mathbf{79.3}$ & $49.8$ \\
    $\mathcal{L}_{\text{instance}}$ & 3D-R18 & $72.7$ & $46.3$ \\
    LSFD (Ours) & 3D-R18 & $77.2$ & $\mathbf{53.7}$ \\
    \bottomrule
    \end{tabular}
\end{center}
\caption{
Comparison to previous methods via finetuning on UCF101 and HMDB51 split 1. 
The first block shows methods with different architectures (Arch) and pretraining datasets, 
while the last two blocks encompass methods with the same architecture and pretraining dataset as our method. $^\dagger$ denotes methods pretrained on UCF101, while $^\ddagger$ denotes methods pretrained on Kinetics-600. The remaining methods are pretrained on Kinetics-400.
}
\label{table:comparison_state_of_the_art}
\end{table}{}

We now evaluate our LSFD method on different downstream tasks.
Previous methods evaluate learned representations based on their performance on action recognition tasks; most commonly, models are finetuned on UCF101 and HMDB51. 
Despite the practical value of finetuning, it is an uncontrolled evaluation~\cite{He:2019:RethinkingINpretraining} and prone to overfitting. Moreover, action classification provides a rather incomplete assessment of the learned representations due to static biases in these datasets \cite{Li:2018:RESOUND}. 
A sparse, global frame sampling strategy, proposed in \cite{Wang:2019:TSN}, works well for action recognition; even in the extreme case where only a single frame is used, accuracy remains high on UCF101. 
This suggests that temporal information is less important for these tasks.
To get a better understanding of our representations, we extend the current evaluation protocol by adding another transfer learning task: action segmentation.

In contrast to action recognition, in which a single action label per video is given, action segmentation uses fine-grained temporal annotations. As the scene and background often stay the same throughout the video, a better temporal understanding is needed in order to temporally segment a video into the different actions that occur.
Additionally, our pretrained model is frozen in this evaluation and serves as a feature extractor -- no finetuning is involved. 

To extend the evaluation via downstream tasks, we analyse the properties of the learned representations more thoroughly.
Here, we are aiming to get a better understanding of which types of attributes are represented by stationary and non-stationary features, and investigate how and why they are different.

We use a 3D-Resnet18 backbone \cite{Hara:2018:3DResnet} in all experiments unless otherwise noted and pool the feature map into a single $512$-dimensional feature vector. We decompose this feature vector into two equal chunks (of size $256$) of stationary and non-stationary features.
We use three separate MLP heads $h_s$, $h_n$, and $h_i$, which are removed after self-supervised training and will not be transferred to downstream tasks. We use a memorybank to store $65.536$ negatives.
We construct long views by sampling $N \cdot L$ frames, which we divide into $N$ non-overlapping sub-sequences of $L$ frames (short views).
We set $L=8$ in all experiments unless otherwise noted, and provide experiments with different values of $N$.
More details regarding the implementation and evaluation are provided in the supplemental material.

\paragraph{Datasets.}
We conduct experiments on four video datasets.
For self-supervised learning, we use videos of \textbf{Kinetics-400}~\cite{Kay:2017:Kinetics} and discard the labels. Our copy of the dataset consists of $234.584$ training and $12.634$ validation videos.
We evaluate the learned representation on \textbf{UCF101}~\cite{Soomro:2012:UCF} and \textbf{HMDB51}~\cite{Kuehne:2011:HMDB} for action recognition and on the \textbf{Breakfast} dataset~\cite{Kuehne:2014:Breakfast} for action segmentation.

\paragraph{Baseline.}
Our most important baseline is a contrastive learning baseline, trained with $\mathcal{L}_{\text{instance}}$ in Eq.~\eqref{eq:vanilla}. This corresponds to traditional contrastive learning~\cite{He:2019:MoCo} applied to videos without an explicit focus on temporal variations.
Here, we use $L=16$ frames and obtain two views by applying augmentations independently.

\subsection{Action Recognition}
The most widely used framework for evaluating self-supervised representations utilizes the self-supervised pretrained weights to initialize a network, and then finetune it on a smaller annotated dataset. We consider the standard benchmarks UCF101 and HMDB51 and compare our method to previous self-supervised video representation methods in Table~\ref{table:comparison_state_of_the_art}. Hyperparameters used for finetuning are provided in the supplementary material.

Since our method is based on RGB-video input only, we exclude multi-modal approaches such as \cite{Piergiovanni:2020:ELo} and \cite{Patrick:2020:GDT}.
The first block in Table~\ref{table:comparison_state_of_the_art} includes methods with shallower networks such as CaffeNet and significantly deeper architectures such as Resnet34 and S3D, and can therefore not be directly compared to our method.
The second and third block includes methods using the same network architecture. 
Here, LSFD is trained with $N=2$ and \Sum\, aggregation.

With the 3D-Resnet18 backbone our method improves over our contrastive baseline ($\mathcal{L}_{\text{instance}}$) on both UCF101 and HMDB51 by a fair margin.
While we outperform previous methods with a similar architecture on HMDB51, our method is inferior to the method of \cite{Jenni:2020:VRL} on UCF101. 
We attribute this smaller relative gain on UCF101 to its inherently static bias \cite{Li:2018:RESOUND, Wang:2019:TSN}, which is less pronounced on HMDB51. 
Moreover, note that since self-supervised learning is more relevant for smaller datasets like HMDB51, the results are more important compared with those on UCF101, which is a mid-sized dataset.
For better comparisons, we additionally provide results with a C3D backbone (second block). Here, we outperform all previous methods. Namely, we outperform  \cite{Jenni:2020:VRL} on UCF101 and HMDB51 by $10\%$ and $12\%$, respectively.

\begin{table}
\begin{center}
    \begin{tabular}{lll|cc}
    \toprule
    \multirow{2}{*}{Loss} & \multirow{2}{*}{Agg} & \multirow{2}{*}{$N$} &\multicolumn{2}{c}{top1 Accuracy}\\
    & & & UCF101 & HMDB51\\
    \midrule
    $\mathcal{L}_{\text{instance}}$ & \Sum & $2$ & $72.7$ & $46.3$ \\
    $\quad+\mathcal{L}_{\text{stationary}}$ & \Sum & $2$ & $74.4$ & $48.7$ \\
    $\quad+\mathcal{L}_{\text{non-stationary}}$ & \Sum & $2$ & $74.8$ & $51.6$ \\
    all & \Sum & $2$ & $77.2$ & $\mathbf{53.7}$ \\
    \midrule
    all & \Linear & $2$ & $77.1$ & $51.3$ \\
    all & \MLP & $2$ & $75.7$ & $49.6$ \\
    all & \GRU & $2$ & $75.5$ & $51.0$ \\
    \midrule
    all & \Sum &$3$ & $77.8$ & $52.1$ \\
    all & \Sum &$4$ & $\mathbf{78.0}$ & $52.3$ \\
    \bottomrule
    \end{tabular}
\end{center}
\caption{
{Ablations via finetuning on UCF101 and HMDB51 for our different loss terms and several design choices, \ie aggregation function (Agg) and number of sub-sequences $N$.}
We see an inverse relation between the aggregation function complexity and performance of the learned representation on the downstream task.
Performance of our method is marginally effected for larger values of $N$ on HMDB51. 
}
\label{table:ablation}
\end{table}{}

\paragraph{How much influence does each loss term have?}
Our LSFD method consists of three separate InfoNCE losses.
We investigate the impact that each of them has on the resulting representation by progressively adding them to $\mathcal{L}_{\text{instance}}$; the results are provided in the first block of Table~\ref{table:ablation}. Adding either $\mathcal{L}_{\text{stationary}}$ or $\mathcal{L}_{\text{non-stationary}}$ improves the performance on both UCF101 and HMDB51, suggesting that both stationary and non-stationary features are useful for action recognition. Note that the relative gain of adding $\mathcal{L}_{\text{non-stationary}}$ is higher on HMDB51 compared with $\mathcal{L}_{\text{stationary}}$. Adding both loss terms gives the highest performance, as is expected.

\paragraph{What is the impact of aggregation?}
\label{sec:aggregation_ablation}
The function used to aggregate non-stationary features in Eq.~\eqref{sim:dynamic} plays a critical role in our proposed method.
The simplest, non-parametric function we consider takes the \Sum\, over non-stationary features. 
We also test parametric and increasingly more complex aggregation functions: a \Linear\, mapping, an \MLP, and a \GRU\, (for details we refer to the supplemental material).
The results are provided in the second block of Table~\ref{table:ablation}.
Overall, we find that the simplest aggregation in form of a \Sum\, yields the highest performance. Using a non-parametric aggregation function puts more load on the backbone, obligating it to do the heavy lifting, whereas parametric aggregation functions relax the task, allowing a potential shortcut via changing the impact of each $\phi_i$. We keep the \Sum\, aggregation for the remaining experiments. 
This observation is in line with learning via equivariances~\cite{Lenc:2015:equivariance}. Essentially, the aggregation function we apply on the feature space can be viewed as a complex transformation of splitting the input data along the temporal axis. A good representation space should translate this complex transformation, \ie mapping the short views to the long views via simple operations such as \Sum. A similar discussion is conducted in~\cite{Counting}.

\paragraph{How many sub-sequences should we use?}
\label{sec:ablation_N}
We ablate the effect of different numbers of sub-sequences $N$ have on the representations in the third block of Table~\ref{table:ablation}.
We observed that training $N=3$ and $N=4$ from scratch was sub-optimal; the task becomes increasingly difficult for larger $N$ (details are provided in the supplemental material). Therefore, we follow a curriculum learning strategy, that uses the pretrained model trained with $N-1$ sub-sequences to initialize training for $N$ sub-sequences.
While increasing $N$ improves the representation on UCF101, we observe a drop on HMDB51.

\subsection{Video Retrieval}
\begin{table*}
\begin{center}
    \begin{tabular}{ll|cccc|cccc}
    \toprule
    \multirow{2}{*}{Method}&\multirow{2}{*}{Architecture}&\multicolumn{4}{c|}{UCF101}&\multicolumn{4}{c}{HMDB51}\\
     & & R@1 & R@5 & R@10 & R@20 & R@1 & R@5 & R@10 & R@20 \\
    \midrule
    VCOP \cite{Xu:2019:VCOP} & R3D & $14.1$ & $30.3$ & $40.4$ & $51.1$ & $7.6$ & $22.9$ & $34.4$ & $48.8$ \\
    VCP \cite{Luo:2020:VCP} & R3D & $18.6$ & $33.6$ & $42.5$ & $53.5$ & $7.6$ & $24.4$ & $36.3$ & $53.6$ \\
    MemDPC \cite{Han:2020:MemDPC} & 2D3D-Resnet18 & $20.2$ & $40.4$ & $52.4$ & $64.7$ & $7.7$ & $25.7$ & $40.6$ & $57.7$ \\
    SpeedNet \cite{Benaim:2020:SpeedNet}  & S3D-G & $13.0$ & $28.1$ & $37.5$ & $49.5$ & -  & - & - & - \\
    PRP \cite{Yao:2020:PRP} & R3D & $22.8$ & $38.5$ & $46.7$ & $55.2$ & $8.2$ & $25.8$ & $38.5$ & $53.3$ \\
    Temp Trans \cite{Jenni:2020:VRL} & 3D-Resnet18 & $26.1$ & $48.5$ & $59.1$ & $69.6$ & - & - & - \\
    CoCLR \cite{Han:2020:CoCLR} & S3D-G & $\mathbf{53.3}$ & $\mathbf{69.4}$ & $\mathbf{76.6}$ & $\mathbf{82.0}$ & $23.2$ & $43.2$ & $53.5$ & $65.5$ \\
    \midrule
    LFSD (Ours) & 3D-Resnet18 & $44.9$ & $64.0$ & $73.2$ & $81.4$ & $\mathbf{26.7}$ & $\mathbf{54.7}$ & $\mathbf{66.4}$ & $\mathbf{76.0}$ \\
    \bottomrule
    \end{tabular}
\end{center}
\caption{
Comparison to other methods via nearest neighbor video retrieval on UCF101 and HMDB51. 
}
\label{table:retrieval}
\end{table*}{}

\label{sec:retrieval}
\begin{figure}
\begin{center}
\includegraphics[width=\linewidth]{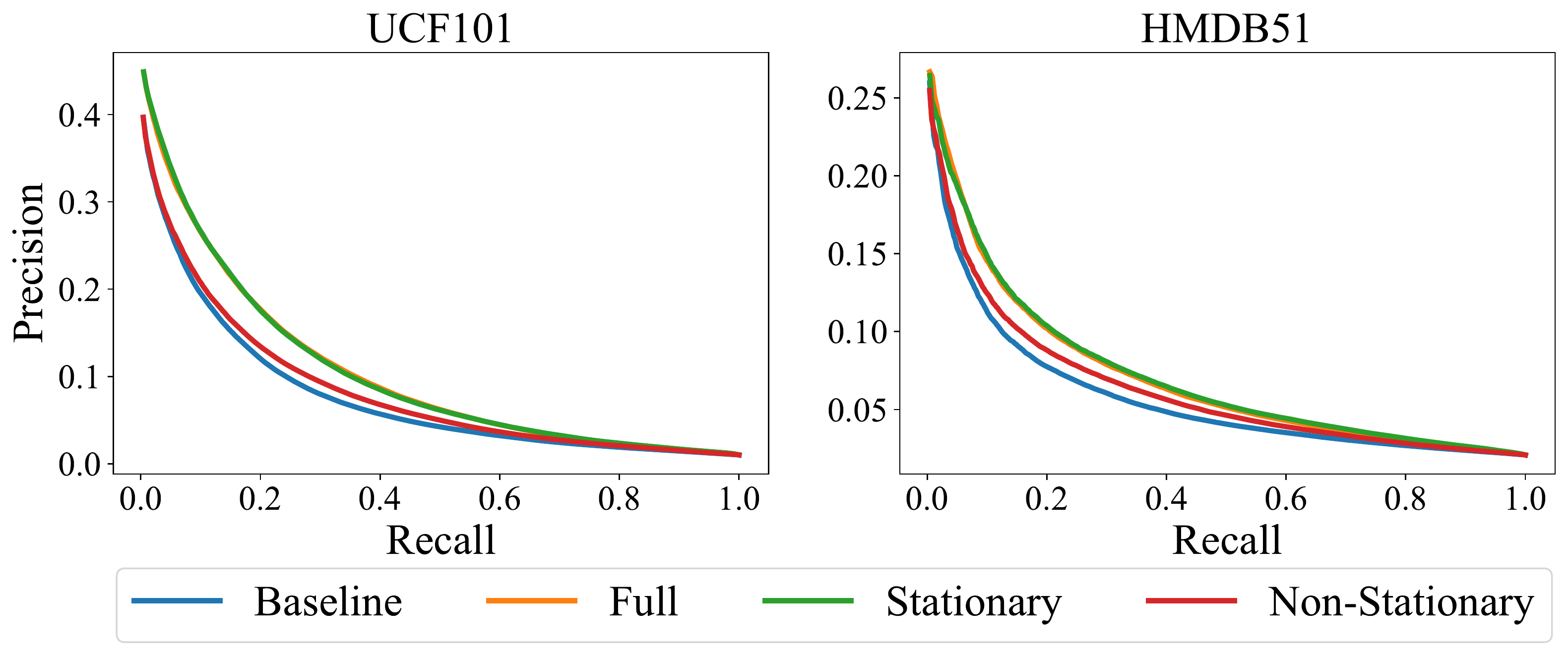}
\end{center}
\caption{Precision-Recall-Curves for UCF101 and HMDB51. Stationary features are superior to non-stationary features on action recognition; both improve over our contrastive baseline.}
\label{fig:precision_recall}
\end{figure}
Next, we evaluate our method on video retrieval.
We follow the protocol of \cite{Xu:2019:VCOP}: We use the pretrained network to extract convolutional features at the last layer for all videos in the training and test set. For each test video we retrieve the top $k$ nearest neighbors from the training videos. For R@$k$ results in Table~\ref{table:retrieval}, we count a correct retrieval if the $k$ nearest neighbors contain at least one video of the same class.

While we improve over previous methods on HMDB51, our retrieval results on UCF101 are inferior to CoCLR~\cite{Han:2020:CoCLR}. Note that CoCLR uses a significantly deeper architecture. This is in line with our observation from the finetuning evaluation, where the relative improvement is higher on HMDB51 compared with UCF101.
Note that this evaluation doesn't measure \textit{precision} of the retrieved samples properly for $k>1$.
For this reason, we also present precision-recall curves in Figure~\ref{fig:precision_recall} (for details see supplementary material). Our stationary features perform on-par with full features; non-stationary features slightly worse. This is even more pronounced on UCF101, where the static bias is higher \cite{Li:2018:RESOUND}.
Both stationary and non-stationary features improve over our contrastive baseline on both datasets.

\subsection{Temporal Action Segmentation}
\begin{figure*}
\begin{center}
\includegraphics[width=\linewidth]{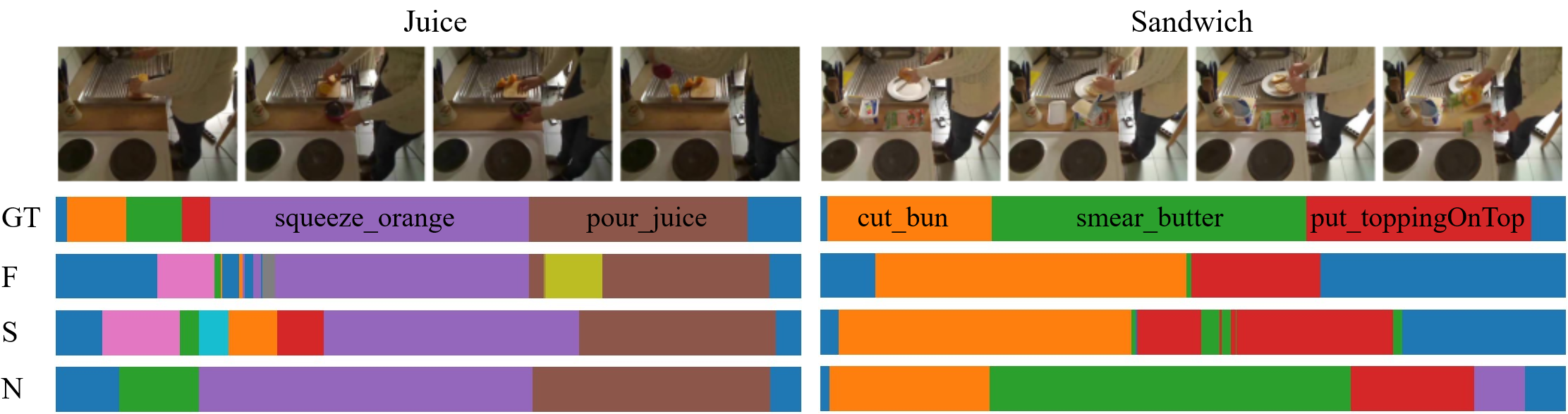}
\end{center}
\caption{Qualitative results of LSFD for two videos from the Breakfast dataset, showing the quality of stationary (S), non-stationary (N), and full (F) features for action segmentation. Non-stationary features provide higher quality representations than stationary features, validating their ability to capture fine-grained temporal variations.}
\label{fig:segmentation}
\end{figure*}
We evaluate the universal representation learning capability of our method via a temporal action segmentation downstream task. Given an untrimmed video, the goal of temporal action segmentation is to simultaneously segment every action in time and classify each obtained segment.
Recent state-of-the-art action segmentation methods such as \cite{Farha:2019:MSTCN, Li:2020:MSTCNpp} train a temporal action segmentation model on top of pre-extracted features of the video frames. 
Usually, video frames are represented using deep 3D CNNs such as I3D~\cite{Carreira:2017:QVA} pretrained on the Kinetics dataset, or hand-crafted features such as improved dense trajectories (IDT)~\cite{IDT}.
In this experiment, we use the Breakfast dataset. This dataset consists of untrimmed videos containing fine-grained actions which are distinguishable mostly via temporal variations in the video, since the scene, actor, and objects remain similar throughout the video.

Using the Breakfast dataset allows us to better evaluate the temporal variation representation capability of our method.
To that end, we use our frozen pretrained model as a feature extractor and compute features for each video frame of the datatset following~\cite{Farha:2019:MSTCN}.
Then, we add a temporal action segmentation model on top, namely MS-TCN~\cite{Farha:2019:MSTCN}, and train for action segmentation in a fully supervised fashion. We use the official publicly available code of MS-TCN for training and evaluation; more training details are provided in the supplemental material.
As the segmentation model relies on the pre-extracted features, this evaluation reveals more reliably than finetuning how well our learned representations are suited for this downstream task that involves a better temporal understanding.
All models in this section are based on a 3D-Resnet18 backbone that operates on RGB input only.

\paragraph{Evaluation metrics.}
For evaluation of the segmentation models, we report the frame-wise accuracy (Acc), segmental edit distance, and the segmental F1 score at overlapping thresholds $10\%$, $25\%$ and $50\%$ as proposed by~\cite{Lea:2017:TCN}.
While the frame-wise accuracy provides a basic rating, it is rather insensitive to over-segmentation errors and short action classes; longer action classes have a higher impact than short action classes.
The segmental edit distance measures how well the model predicts the ordering of action classes, and is not impacted by the duration of action classes.
The segmental F1 score measures the general quality of the segmentation model, as it penalizes over-segmentation and is also insensitive to the duration of the action classes.
More specifically, we determine for each predicted action segment whether it is a true or false positive by taking a threshold on the IoU with the ground truth. Then we compute precision and recall summed over all classes and compute $\text{F1}=2\frac{prec\cdot recall}{prec + recall}$.

\begin{table}
\begin{center}
    \begin{tabular}{l|ccccc}
    \toprule
    Method & Acc & Edit & \multicolumn{3}{c}{F1@\{10, 25, 50\}}\\
    \midrule
    Random init. & $32.6$ & $41.2$ & $39.3$ & $32.4$ & $21.8$ \\
    Kinetics sup. & $45.1$ & $54.5$ & $47.1$ & $41.7$ & $31.0$ \\
    \midrule
    $\mathcal{L}_{\text{instance}}$ & $57.9$ & $50.3$ & $44.6$ & $39.9$ & $31.4$ \\
    LSFD, F & $60.1$ & $56.3$ & $46.1$ & $41.7$ & $32.6$ \\
    LSFD, S & $58.9$ & $54.7$ & $40.6$ & $35.7$ & $28.8$ \\
    LSFD, N & $\mathbf{60.6}$ & $\mathbf{60.0}$ & $\mathbf{52.0}$ & $\mathbf{42.8}$ & $\mathbf{35.3}$ \\
    \bottomrule
    \end{tabular}
\end{center}
\caption{
Action Segmentation on the Breakfast dataset split 1.
We report results for several baselines and our LSFD method. We further split up the full (F) feature in stationary (S) and non-stationary (N) features to investigate their effect. The backbone of all of the feature extraction models is 3D-ResNet18.
}
\label{table:segmentation}
\end{table}{}

\paragraph{Results.}
In Table~\ref{table:segmentation} we provide the results for our unsupervised LSFD method as well as several baselines, including random initialization, full Kinetics supervision, and our contrastive baseline ($\mathcal{L}_{\text{instance}}$). 
Interestingly, we observe that all features obtained via unsupervised pretraining, \ie the second block of the table, improve over a model trained with Kinetics supervision. This validates our argument that unsupervised learning may provide richer representations, capable of transferring better to different, unseen tasks.
Furthermore, our method improves over the contrastive baseline as well as the supervised baseline by a significant margin, demonstrating that our long and short views enable a better temporal understanding.
While the difference in accuracy is noticeable, it is even more prominent in the segmental edit distance and F1 scores, which better measure the overall quality of the segmentations.

To investigate the feature decomposition we use only the stationary or only the non-stationary features of our pretrained model as input to the segmentation model.
The quantitative results in Table~\ref{table:segmentation} show that our non-stationary features outperform stationary features across all metrics, providing higher quality representations for the temporal segmentation model, and are consequently better suited for fine-grained temporal action segmentation.
As can be seen in the raw frames in Figure~\ref{fig:segmentation}, the static information across the video remains similar for most of the temporal segments. Hence, temporally segmenting such videos to fine-grained actions is a challenging problem requiring high quality features in terms of temporal variation representation. Moreover, the difference between the temporal segmentation performance of our stationary and non-stationary features also validates the feature space decomposition capability of our method.
Additionally, we provide some qualitative results in Figure~\ref{fig:segmentation}, where we observe a higher quality of non-stationary features compared with stationary features.
As mentioned before, Breakfast is a challenging temporal action segmentation dataset due to the similarity of the objects, actor, and scene throughout the entire video. 
Therefore, the non-stationary features, which better represent the temporal variations in the video, empower the action segmentation model.

\begin{figure*}[t]
    \begin{minipage}[t]{1\textwidth}
    \centering
    \includegraphics[width=.5\textwidth]{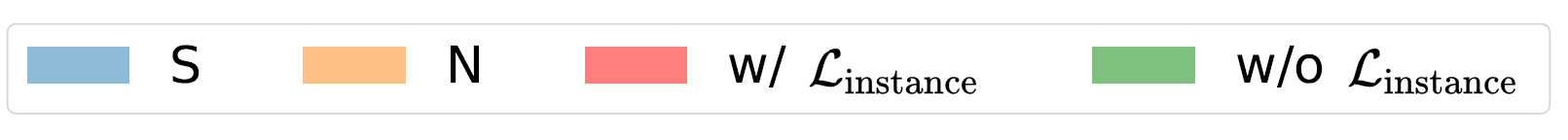}
    
    \begin{minipage}[t]{.22\textwidth}
    \centering
    \small
    \includegraphics[width=1\textwidth]{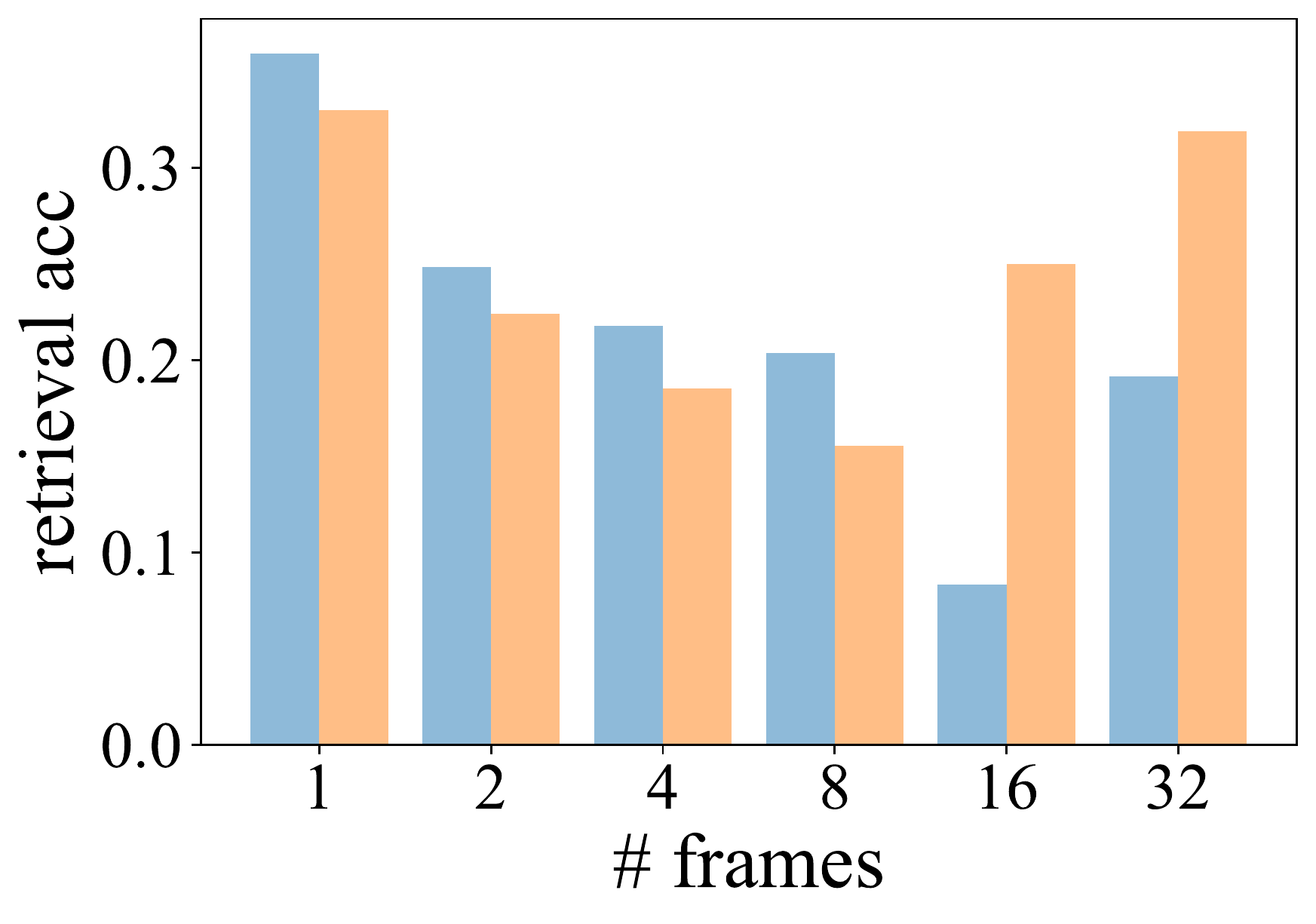}
    
    (a)
    \end{minipage}
    \begin{minipage}[t]{.22\textwidth}
    \centering
    \small
    \includegraphics[width=1\textwidth]{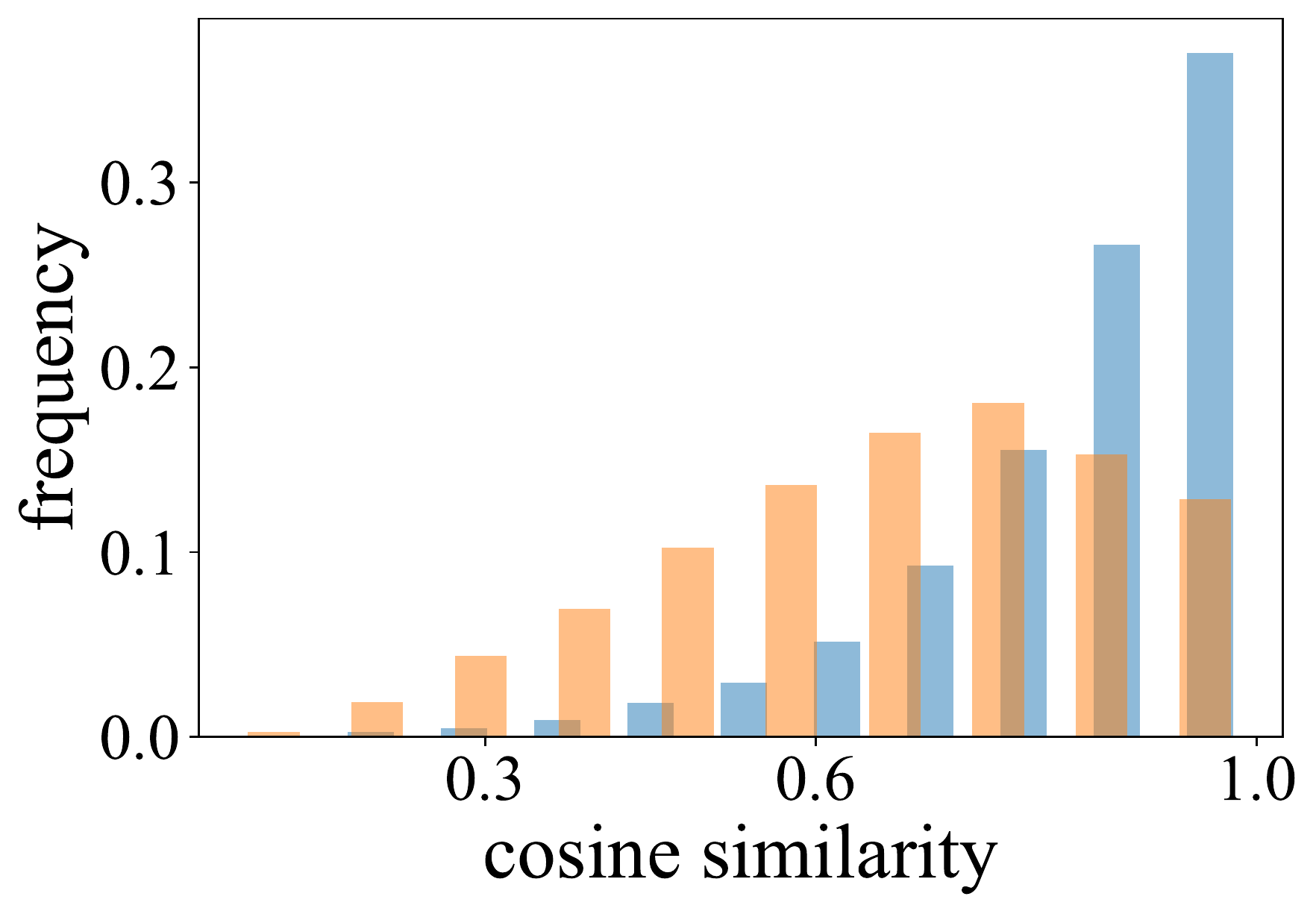}
    
    (b)
    \end{minipage}
    \begin{minipage}[t]{.22\textwidth}
    \centering
    \small
    \includegraphics[width=1\textwidth]{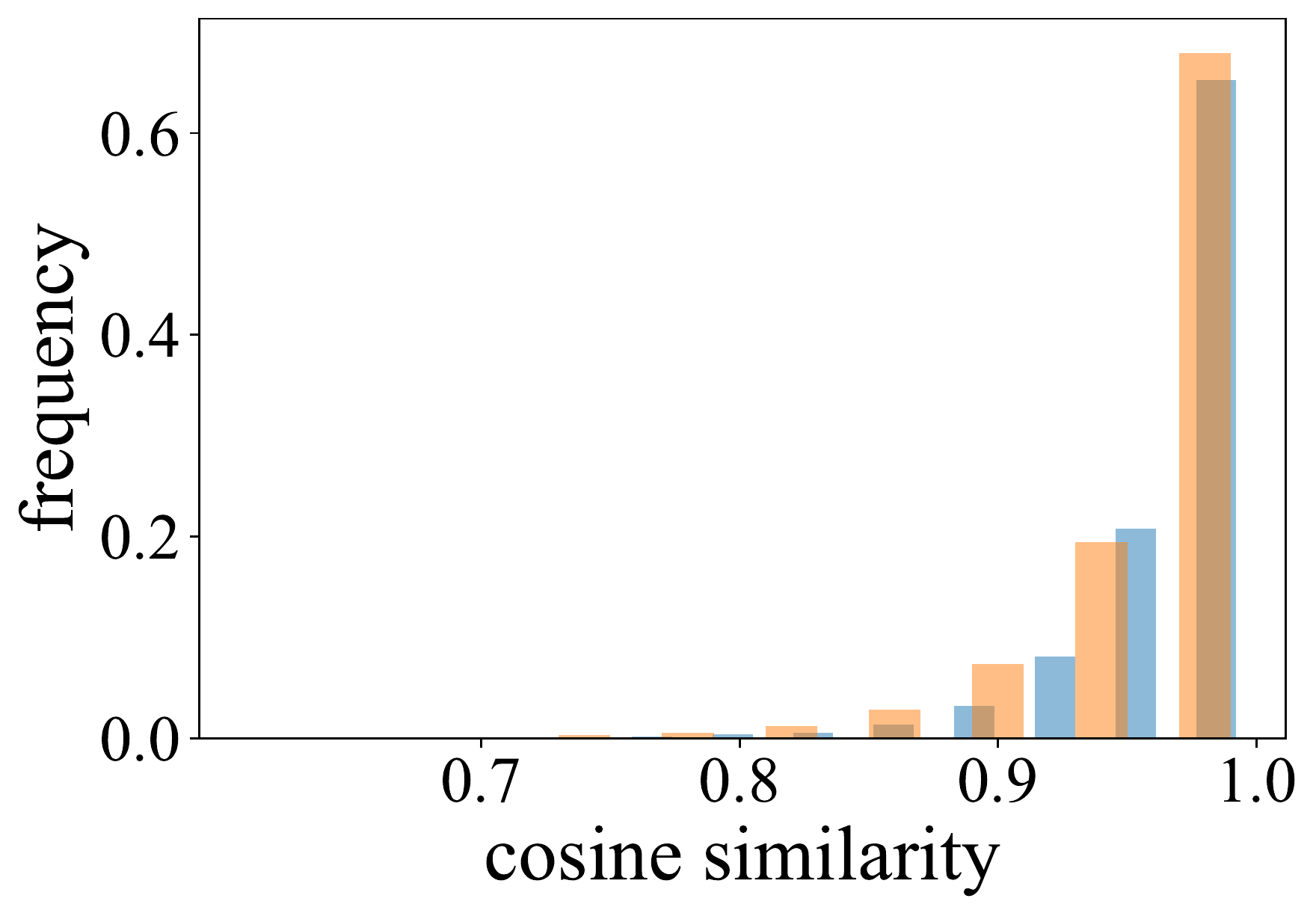}    
    
    (c)
    \end{minipage}
    \begin{minipage}[t]{.22\textwidth}
    \centering
    \small
    \includegraphics[width=1\textwidth]{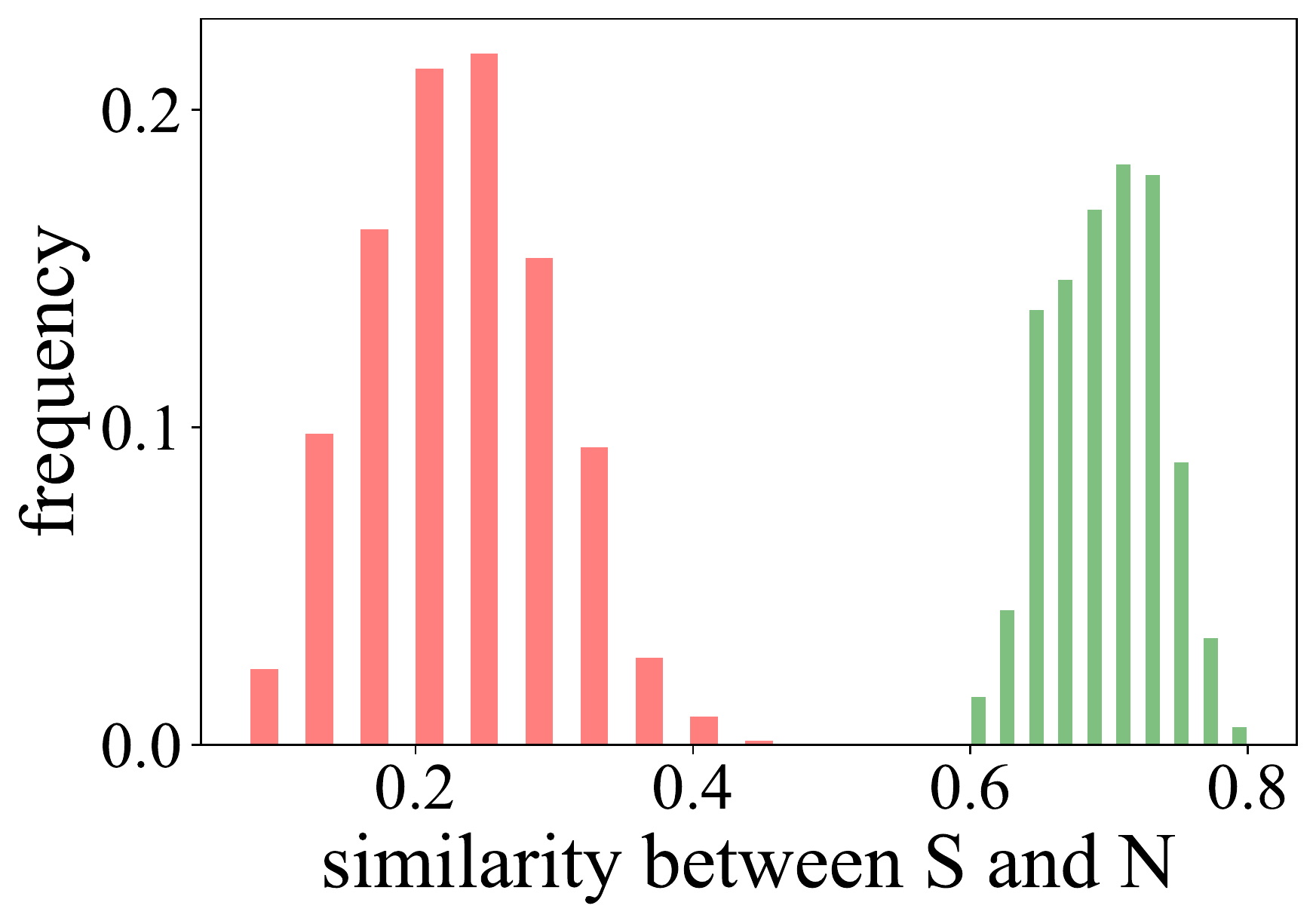}
    
    (d)
    \end{minipage}
    \end{minipage}
    \caption{\textbf{Feature Decomposition Analyses.} (a) Retrieval accuracy of different subsets of HMDB51 based on the number of frames needed for classification. We observe that non-stationary features (N) outperform stationary features (S) on retrieving videos that require more frames (long views). The opposite happens for videos that require less frames (short views).  Cosine similarities over time (b) with and (c) without $\mathcal{L}_{\text{instance}}$. We compute similarities of S and N features over time and show the histogram of the computed similarities on HMDB51. (b) S features are centered around high values when training with $\mathcal{L}_{\text{instance}}$; N features are distributed more uniformly. (c) When training without $\mathcal{L}_{\text{instance}}$ both S and N features are stable over time. (d) Similarities between S and N features when trained with and without $\mathcal{L}_{\text{instance}}$. We observe that removing $\mathcal{L}_{\text{instance}}$ results in a degenerate solution where S and N features are similar.}
\label{fig:feature_decomposition_analyses}
\end{figure*}

\subsection{Feature Decomposition Analyses}
The experiments above validate our hypothesis on feature decomposition by evaluating the performance of stationary and non-stationary features on dedicated downstream tasks. In the following, we conduct more specific analyses to obtain a better insight into our decomposed features.
For experimental details we refer to the supplemental material.

\paragraph{Do we capture short- and long-term attributes?}
Figure~\ref{fig:feature_decomposition_analyses}(a) shows the retrieval accuracy of stationary and non-stationary features among the videos that can be classified with different numbers of frames.
In each case, we exclude videos that are correctly classified with fewer frames, \ie video that can be classified with a single frame are excluded from the set of videos that can be classified with two frames, etc. While stationary features achieve stronger retrieval accuracy on videos that can be classified with less frames (short views), \ie less temporal context, non-stationary features are more beneficial for videos that require more frames (long views) and a longer temporal context.

\paragraph{Are stationary features more stable over time?}
We divide a given video into clips of $16$ frames, compute the stationary and non-stationary features of all clips and similarities of the features over time. We show the histogram of the computed similarities on HMDB51 in Figure~\ref{fig:feature_decomposition_analyses}(b). Stationary features are centered around high values, while non-stationary features are distributed more uniformly. This suggests that stationary features remain more stable over time, whereas non-stationary feature vary.

\paragraph{What is the impact of $\mathcal{L}_{\text{instance}}$?}
Figure~\ref{fig:feature_decomposition_analyses}(c) shows the same histogram as Figure~\ref{fig:feature_decomposition_analyses}(b) for a network trained without $\mathcal{L}_{\text{instance}}$. We observe that in contrast to Figure~\ref{fig:feature_decomposition_analyses}(b), stationary and non-stationary features behave similarly -- both are stable over time. Moreover, we compare the histogram of similarities between stationary and non-stationary features of long views with and without $\mathcal{L}_{\text{instance}}$ in Figure~\ref{fig:feature_decomposition_analyses}(d). Stationary and non-stationary features are very similar to each other when trained without $\mathcal{L}_{\text{instance}}$, suggesting a degenerate solution that copies stationary features as non-stationary ones. One reason why $\mathcal{L}_{\text{instance}}$ avoids this could be that it pushes the network towards exploiting the full capacity of the feature space, preventing redundant information in the full feature space.

\section{Conclusion}
In this paper, we have introduced a novel method to decompose video representations into stationary and non-stationary features via contrastive learning from long and short views.
We evaluated the learned representations extensively on multiple downstream tasks and datasets, and investigated various design choices and the role of stationary and non-stationary features.
Overall, we find an interesting correlation between the type of feature and the nature of the downstream task: Stationary features perform better on tasks and datasets with static biases, such as action recognition on UCF101, while non-stationary features are more beneficial for action segmentation, which requires better temporal understanding.
We demonstrated a substantial gain in performance on the HMDB51 dataset for action recognition, and outperform a supervised baseline on the Breakfast dataset for action segmentation.
We are the first to conduct an evaluation for video representation learning via action segmentation, and there are two main takeaways:
First, action segmentation provides an interesting and challenging downstream task to evaluate the quality of learned representations.
Second, unsupervised video representation learning has the potential to significantly improve representations used for action segmentation, and provides a promising approach for future research. We have found that a substantial improvement over supervised pretraining can be achieved. 
Furthermore, we analysed our feature decomposition and found that stationary features are more stable over time, while non-stationary features vary.

{\footnotesize JG has been supported by the Deutsche Forschungsgemeinschaft (DFG, German Research Foundation) GA1927/4-2 (FOR 2535 Anticipating Human Behavior) and the ERC Starting Grant ARCA (677650).}

{\small
\bibliographystyle{ieee_fullname}
\bibliography{main}

\begin{thebibliography}{10}\itemsep=-1pt

\bibitem{BidirectionalFeaturePrediction}
Nadine Behrmann, Juergen Gall, and Mehdi Noroozi.
\newblock Unsupervised video representation learning by bidirectional feature
  prediction.
\newblock In {\em WACV}, 2021.

\bibitem{Benaim:2020:SpeedNet}
Sagie Benaim, Ariel Ephrat, Oran Lang, Inbar Mosseri, William~T. Freeman,
  Michael Rubinstein, Michal Irani, and Tali Dekel.
\newblock Speednet: Learning the speediness in videos.
\newblock In {\em CVPR}, 2020.

\bibitem{Carreira:2017:QVA}
Jo{\~{a}}o Carreira and Andrew Zisserman.
\newblock Quo vadis, action recognition? {A} new model and the kinetics
  dataset.
\newblock In {\em CVPR}, 2017.

\bibitem{Chen:2020:SimCLR}
Ting Chen, Simon Kornblith, Mohammad Norouzi, and Geoffrey Hinton.
\newblock A simple framework for contrastive learning of visual
  representations.
\newblock In {\em ICML}, 2020.

\bibitem{Chen:2020:SimSiam}
Xinlei Chen and Kaiming He.
\newblock Exploring simple siamese representation learning.
\newblock {\em arXiv}, abs/2011.10566, 2020.

\bibitem{Dosovitskiy:2016:DUF}
Alexey Dosovitskiy, Philipp Fischer, Jost~Tobias Springenberg, Martin
  Riedmiller, and Thomas Brox.
\newblock Discriminative unsupervised feature learning with exemplar
  convolutional neural networks.
\newblock In {\em TPAMI}, 2016.

\bibitem{Farha:2019:MSTCN}
Yazan~Abu Farha and Juergen Gall.
\newblock {MS}-{TCN}: Multi-stage temporal convolutional network for action
  segmentation.
\newblock In {\em CVPR}, 2019.

\bibitem{Grill:2020:BYOL}
Jean-Bastien Grill, Florian Strub, Florent Altch\'{e}, Corentin Tallec, Pierre
  Richemond, Elena Buchatskaya, Carl Doersch, Bernardo Avila~Pires, Zhaohan
  Guo, Mohammad Gheshlaghi~Azar, Bilal Piot, Koray Kavukcuoglu, R\'{e}mi Munos,
  and Michal Valko.
\newblock Bootstrap your own latent - a new approach to self-supervised
  learning.
\newblock In {\em NeurIPS}, 2020.

\bibitem{Han:2019:DPC}
Tengda Han, Weidi Xie, and Andrew Zisserman.
\newblock Video representation learning by dense predictive coding.
\newblock In {\em ICCV Workshop}, 2019.

\bibitem{Han:2020:MemDPC}
Tengda Han, Weidi Xie, and Andrew Zisserman.
\newblock Memory-augmented dense predictive coding for video representation
  learning.
\newblock In {\em ECCV}, 2020.

\bibitem{Han:2020:CoCLR}
Tengda Han, Weidi Xie, and Andrew Zisserman.
\newblock Self-supervised co-training for video representation learning.
\newblock In {\em NeurIPS}, 2020.

\bibitem{Hara:2018:3DResnet}
Kensho Hara, Hirokatsu Kataoka, and Yutaka Satoh.
\newblock Can spatiotemporal {3D CNNs} retrace the history of {2D CNNs} and
  {ImageNet}?
\newblock In {\em CVPR}, 2018.

\bibitem{He:2019:MoCo}
Kaiming He, Haoqi Fan, Yuxin Wu, Saining Xie, and Ross Girshick.
\newblock Momentum contrast for unsupervised visual representation learning.
\newblock In {\em CVPR}, 2020.

\bibitem{He:2019:RethinkingINpretraining}
Kaiming He, Ross Girshick, and Piotr Dollar.
\newblock Rethinking imagenet pre-training.
\newblock In {\em ICCV}, 2019.

\bibitem{Jenni:2020:VRL}
Simon Jenni, Givi Meishvili, and Paolo Favaro.
\newblock Video representation learning by recognizing temporal
  transformations.
\newblock In {\em ECCV}, 2020.

\bibitem{Jing:2018:SSS}
Longlong Jing and Yingli Tian.
\newblock Self-supervised spatiotemporal feature learning via video rotation
  prediction.
\newblock {\em arXiv}, abs/1811.11387, 2018.

\bibitem{Kay:2017:Kinetics}
Will Kay, Joao Carreira, Karen Simonyan, Brian Zhang, Chloe Hillier, Sudheendra
  Vijayanarasimhan, Fabio Viola, Tim Green, Trevor Back, Paul Natsev, Mustafa
  Suleyman, and Andrew Zisserman.
\newblock The kinetics human action video dataset.
\newblock {\em arXiv}, abs/1705.06950, 2017.

\bibitem{Kim:2019:STC}
Dahun Kim, Donghyeon Cho, and In~So Kweon.
\newblock Self-supervised video representation learning with space-time cubic
  puzzles.
\newblock In {\em AAAI}, 2019.

\bibitem{Kingma:2015:Adam}
Diederik Kingma and Jimmy Ba.
\newblock Adam: {A} method for stochastic optimization.
\newblock In {\em ICLR}, 2015.

\bibitem{Kuehne:2014:Breakfast}
Hilde Kuehne, Ali~B. Arslan, and Thomas Serre.
\newblock The language of actions: Recovering the syntax and semantics of
  goal-directed human activities.
\newblock In {\em CVPR}, 2014.

\bibitem{Kuehne:2011:HMDB}
Hilde Kuehne, Hueihan Jhuang, Est{\'i}baliz Garrote, Tomaso Poggio, and Thomas
  Serre.
\newblock {HMDB}: A large video database for human motion recognition.
\newblock In {\em ICCV}, 2011.

\bibitem{Lea:2017:TCN}
Colin {Lea}, Michael~D. {Flynn}, Ren\'{e} {Vidal}, Austin {Reiter}, and
  Gregory~D. {Hager}.
\newblock Temporal convolutional networks for action segmentation and
  detection.
\newblock In {\em CVPR}, 2017.

\bibitem{Lee:2017:URL}
Hsin{-}Ying Lee, Jia{-}Bin Huang, Maneesh Singh, and Ming{-}Hsuan Yang.
\newblock Unsupervised representation learning by sorting sequences.
\newblock In {\em ICCV}, 2017.

\bibitem{Lenc:2015:equivariance}
Karel {Lenc} and Andrea {Vedaldi}.
\newblock Understanding image representations by measuring their equivariance
  and equivalence.
\newblock In {\em CVPR}, 2015.

\bibitem{Li:2020:MSTCNpp}
Shi-Jie Li, Yazan AbuFarha, Yun Liu, Ming-Ming Cheng, and Juergen Gall.
\newblock {MS}-{TCN}++: Multi-stage temporal convolutional network for action
  segmentation.
\newblock {\em TPAMI}, 2020.

\bibitem{Li:2018:RESOUND}
Yingwei Li, Yi Li, and Nuno Vasconcelos.
\newblock Resound: Towards action recognition without representation bias.
\newblock In {\em ECCV}, 2018.

\bibitem{Luo:2020:VCP}
Dezhao Luo, Chang Liu, Yu Zhou, Dongbao Yang, Can Ma, Qixiang Ye, and Weiping
  Wang.
\newblock Video cloze procedure for self-supervised spatio-temporal learning.
\newblock In {\em AAAI}, 2020.

\bibitem{Misra:2019:PIRL}
Ishan Misra and Laurens van~der Maaten.
\newblock Self-supervised learning of pretext-invariant representations.
\newblock In {\em CVPR}, 2020.

\bibitem{Misra:2016:SAL}
Ishan Misra, C.~Lawrence Zitnick, and Martial Hebert.
\newblock Shuffle and learn: Unsupervised learning using temporal order
  verification.
\newblock In {\em ECCV}, 2016.

\bibitem{Counting}
Mehdi Noroozi, Hamed Pirsiavash, and Paolo Favaro.
\newblock Representation learning by learning to count.
\newblock In {\em ICCV}, 2017.

\bibitem{Patrick:2020:GDT}
Mandela Patrick, Yuki~M. Asano, Polina Kuznetsova, Ruth Fong, João~F.
  Henriques, Geoffrey Zweig, and Andrea Vedaldi.
\newblock Multi-modal self-supervision from generalized data transformations.
\newblock {\em arXiv}, abs/2003.04298, 2020.

\bibitem{Piergiovanni:2020:ELo}
A.~J. Piergiovanni, Anelia Angelova, and Michael~S. Ryoo.
\newblock Evolving losses for unsupervised video representation learning.
\newblock In {\em CVPR}, 2020.

\bibitem{Purushwalkam:2020:DCS}
Senthil Purushwalkam and Abhinav Gupta.
\newblock Demystifying contrastive self-supervised learning: Invariances,
  augmentations and dataset biases.
\newblock In {\em NeurIPS}, 2020.

\bibitem{Sohn:2016:Npair}
Kihyuk Sohn.
\newblock Improved deep metric learning with multi-class n-pair loss objective.
\newblock In {\em NIPS}, 2016.

\bibitem{Soomro:2012:UCF}
Khurram Soomro, Amir~Roshan Zamir, and Mubarak Shah.
\newblock Ucf101: A dataset of 101 human actions classes from videos in the
  wild.
\newblock {\em arXiv}, abs/1212.0402, 2012.

\bibitem{Sun:2019:CBT}
Chen Sun, Fabien Baradel, Kevin Murphy, and Cordelia Schmid.
\newblock Learning video representations using contrastive bidirectional
  transformer.
\newblock {\em arXiv}, abs/1906.05743, 2019.

\bibitem{Tian:2019:CMC}
Yonglong Tian, Dilip Krishnan, and Phillip Isola.
\newblock Contrastive multiview coding.
\newblock {\em arXiv}, abs/1906.05849, 2019.

\bibitem{Tokmakov:2020:LA_IDT}
Pavel Tokmakov, Martial Hebert, and Cordelia Schmid.
\newblock Unsupervised learning of video representations via dense trajectory
  clustering.
\newblock In {\em ECCV Workshops}, 2020.

\bibitem{Oord:2018:CPC}
A{\"a}ron van~den Oord, Yazhe Li, and Oriol Vinyals.
\newblock Representation learning with contrastive predictive coding.
\newblock {\em arXiv}, abs/1807.03748, 2018.

\bibitem{IDT}
Heng {Wang} and Cordelia {Schmid}.
\newblock Action recognition with improved trajectories.
\newblock In {\em ICCV}, 2013.

\bibitem{Wang:2019:MA}
Jiangliu Wang, Jianbo Jiao, Linchao Bao, Shengfeng He, Yunhui Liu, and Wei Liu.
\newblock Self-supervised spatio-temporal representation learning for videos by
  predicting motion and appearance statistics.
\newblock In {\em CVPR}, 2019.

\bibitem{Wang:2020:PP}
Jiangliu Wang, Jianbo Jiao, and Yun-Hui Liu.
\newblock Self-supervised video representation learning by pace prediction.
\newblock In {\em ECCV}, 2020.

\bibitem{Wang:2019:TSN}
L. {Wang}, Y. {Xiong}, Z. {Wang}, Y. {Qiao}, D. {Lin}, X. {Tang}, and L. {Van
  Gool}.
\newblock Temporal segment networks for action recognition in videos.
\newblock {\em TPAMI}, 2019.

\bibitem{Xu:2019:VCOP}
Dejing Xu, Jun Xiao, Zhou Zhao, Jian Shao, Di Xie, and Yueting Zhuang.
\newblock Self-supervised spatiotemporal learning via video clip order
  prediction.
\newblock In {\em CVPR}, 2019.

\bibitem{Yao:2020:PRP}
Yuan Yao, Chang Liu, Dezhao Luo, Yu Zhou, and Qixiang Ye.
\newblock Video playback rate perception for self-supervised spatio-temporal
  representation learning.
\newblock In {\em CVPR}, 2020.

\bibitem{Zhuang:2020:VIE}
Chengxu Zhuang, Tianwei She, Alex Andonian, Max~Sobol Mark, and Daniel Yamins.
\newblock Unsupervised learning from video with deep neural embeddings.
\newblock In {\em CVPR}, 2020.

\end{thebibliography}
}
\section*{Supplemental Material}

\paragraph{Architecture.}
We use a 3D-Resnet18 backbone \cite{Hara:2018:3DResnet} unless otherwise noted and pool the feature map into a single $512$-dimensional feature vector.
The first $256$ neurons serve as the static feature; the last $256$ neurons serve as the dynamic feature.
The MLP head $h_{i}$ has $512$ hidden units with ReLU activation, $h_{s}$ and $h_{n}$ have $256$ hidden units. Note that the MLP heads are removed after self-supervised training and will not be transferred to downstream tasks.

In Section~4.1 we investigate the influence of different aggregation functions, namely \Sum, \Linear, \MLP, and \GRU.
The \Sum\, simply takes the sum over the non-stationary features of the short views, \ie $\sum_{i=1}^N \phi_s^{(i)}$.
For \Linear\, and \MLP\, we first concatenate the non-stationary features, \ie $\left(\phi_s^{(1)}, \dots, \phi_s^{(N)}\right)\in\mathbb{R}^{N\times 256}$, and then apply a linear layer or an MLP, respectively, mapping from $\mathbb{R}^{N\times 256}$ to $\mathbb{R}^{256}$. The \MLP\, has $N\times 256$ hidden units with ReLU activation.
The \GRU\, aggregates the sequence of non-stationary features $\left(\phi_s^{(i)}\right)_{i=1}^N$ to produce a single aggregated feature $\phi_g \in \mathbb{R}^{256}$ of the same dimension as the non-stationary features of the long view. We use a one-layer ConvGRU with a kernel size of $1$.

\paragraph{Datasets.}
We conduct experiments on four video datasets.
For self-supervised learning, we use videos of \textbf{Kinetics-400}~\cite{Kay:2017:Kinetics} and discard the labels. Our copy of the dataset consists of $234.584$ training and $12.634$ validation videos.
We evaluate the learned representation on \textbf{UCF101}~\cite{Soomro:2012:UCF} and \textbf{HMDB51}~\cite{Kuehne:2011:HMDB} for action recognition and on the \textbf{Breakfast} dataset~\cite{Kuehne:2014:Breakfast} for action segmentation.

\paragraph{Implementation Details.}
For self-supervised pretraining, we use the Adam optimizer \cite{Kingma:2015:Adam} with weight decay $1e-5$, a batch size of $128$ and an initial learning rate of $1e-3$, that is decreased by a factor of $10$ when the validation loss plateaus.
We set $\tau=0.1$ and $m=0.99$ for the momentum update of the key encoder.
We use a memory bank size of $65.536$ as in \cite{He:2019:MoCo}.

For finetuning we sample clips of $16$ frames with a temporal stride of $3$, and train the model end-to-end using the standard cross entropy loss.
We train for $500$ epochs using the Adam optimizer with an initial learning rate of $1e-4$, which we reduce at epoch $300$ and $400$ by a factor of $10$.
Weight decay is set to $1e-5$ and we use a dropout of $0.9$.
During inference we sample $10$ clips from each test video, and use ten crop. The predictions are averaged to produce the final prediction of each test video.

\paragraph{Views and Augmentations.}
We construct long views by sampling $N \cdot L$ frames with a temporal stride of $3$. We set $L=8$ in all experiments, unless otherwise noted and provide experiments with different values of $N$.
Given a long view of $N \cdot L$ frames, we divide them into $N$ non-overlapping sub-sequences of $L$ frames, which serve as short views. We apply spatial augmentations, such as crop and horizontal flip, and color augmentations to each view independently. 
More specifically, we use random resized crop with probability $p=1.0$, where a spatial patch is selected covering $50\%$ to $100\%$ of the original frame with an aspect ratio between $3/4$ and $4/3$. Then, we resize the patch to the size $128\times 128$.
Horizontal flip is applied with probability $p=0.5$.
For color augmentations we use random color drop with probability $p=0.1$, and apply color jitter with probability $p=1.0$, where brightness, contrast, saturation and hue are shifted. We use a maximum brightness adjustment of $0.5$, contrast of $0.5$, saturation of $0.5$, and hue of $0.25$.
The different views of a video sequence (long and short views) are augmented independently, but within a single view, the frames are augmented consistently, \ie the same crop, color augmentation, etc. is selected for all frames of this view. 
During finetuning we keep the random crop and horizontal flip, but only apply color jitter as described above with a probability of $p=0.3$.

\paragraph{Evaluation.}
To evaluate the learned video representations, we follow the most widely adopted approach of \textit{finetuning}: We use the pretrained weights to initialize the 3D-Resnet18 backbone network, add a randomly initialized linear layer for classification and then train it end-to-end on split 1 of UCF101 and HMDB51. 

The exact choice of the framework used for finetuning influences the final accuracies substantially; an apples-to-apples comparison between different methods is impossible.
For this reason, we additionally provide retrieval results. Here, the pretrained network serves as a feature extractor and is kept fixed.
We extract features for all videos in the dataset and compute R@$k$: For each video in the test set we retrieve the top $k$ nearest neighbor and count a correct retrieval if at least one of the videos is of the same class as the test video. 
Note that R@$k$ does not measure the precision of the retrieved results. Therefore, we also present precision-recall-curves. Here, we compute precision and recall for all values of $k$ and plot the resulting curves. Precision and recall are calculated as it is the standard approach in retrieval. Precision is the fraction of relevant instances among the retrieved instances, while recall is the fraction of relevant instances that were retrieved.

Finally, we evaluate our models on another transfer learning task: action segmentation. We use the pretrained model to extract features from the video frames of the Breakfast dataset and subsequently train a temporal action segmentation model on top of them. Again, this evaluation does not involve any finetuning on the target dataset.
This provides a more elaborate assessment of the learned representations.
We evaluate the segmentation model via frame-wise accuracy, segmental edit distance and F1 scores at overlapping thresholds $10\%$, $25\%$ and $50\%$.
More specifically, for the F1 scores we determine for each predicted action segment whether it is a true or false positive by taking a threshold on the IoU with the ground truth. Then we compute precision and recall summed over all classes and compute $\text{F1}=2\frac{prec\cdot recall}{prec + recall}$.

\paragraph{Curriculum learning for larger $N$.}
We investigate the effect different numbers of sub-sequences have on the learned representations. We keep $L=8$ fixed in this experiment and only vary $N$.
We notice that training LSFD with $N>2$ from scratch is sub-optimal, decreasing the performance on both UCF101 and HMDB51, see Table~\ref{table:ablation_N}. We attribute this to the increased difficulty of the task for larger $N$, and propose to follow a curriculum learning strategy instead. Here, we use the pretrained weights obtained from training LSFD with $N-1$ to initialize the training for $N$. We train $N=2$ from scratch for $100$ epochs, and subsequently train $N=3$ and $N=4$ for $40$ epochs with a reduced learning rate and weight decay of $1e-4$ and $1e-6$, respectively.
As evident in Table~\ref{table:ablation_N}, this approach improves the downstream performance compared with training from scratch.

\begin{table}
\begin{center}
    \begin{tabular}{ll|cc}
    \toprule
    \multirow{2}{*}{$N$}&\multirow{2}{*}{Training}&\multicolumn{2}{c}{top1 Accuracy}\\
     & & UCF101 & HMDB51\\
    \midrule
    $2$ & scratch & $77.2$ & $53.7$ \\
    $3$ & scratch & $75.5$ & $49.6$ \\
    $4$ & scratch & $76.5$ & $50.9$ \\
    $3$ & curriculum & $77.8$ & $52.1$ \\
    $4$ & curriculum & $78.0$ & $52.3$ \\
    \bottomrule
    \end{tabular}
\end{center}
\caption{
{Finetuning results on UCF101 and HMDB51. We train LSFD using different numbers of sub-sequences $N$, either from scratch (scratch) or in a curriculum learning regime (curriculum). We notice that training from scratch is sub-optimal, compared to curriculum learning, which we attribute to the increased difficulty of the task for larger $N$.}
}
\label{table:ablation_N}
\end{table}{}

\paragraph{Implementation details for MS-TCN.}
We use the official publicly available code of MS-TCN~\cite{Farha:2019:MSTCN} for training and evaluation.
The MS-TCN model consists of four stages, each containing ten dilated convolutional layers.
We train the model for $295$ epochs using the Adam optimizer with an initial learning rate of $0.0005$ and the \texttt{ReduceLROnPlateau} learning rate scheduler on the average loss per epoch.
The first layer of MS-TCN adjusts the dimension of the input features (\ie $512$ for full features and $256$ for stationary and non-stationary features in our experiments) using a $1 \times 1$ convolution; the remaining layers have $64$ channels.

\paragraph{Feature Decomposition Analyses.}
In this section, we are aiming to get a better understanding of our feature decomposition; specifically, we are interested in the difference between our stationary and non-stationary features. 
To that end, we use the pretrained model as a feature extractor (without the MLP heads $h_s$ and $h_n$). To compute similarities between two feature vectors $x$ and $y$, we use the cosine similarity:
\begin{equation*}
    \frac{x^T y}{\|x\|\|y\|}.
\end{equation*}

Furthermore, we compute retrieval accuracies among videos that can be classified with different numbers of frames. 
First, we train separate model receiving $N=1,2,4,8,16,32$ frames as input.
Then, we group the videos into disjoint subsets based on the numbers of frames that are needed for classification. For $N=1$ this subset consists of all videos that are correctly classified by the model trained with $N=1$ frames. For $N=2$ the subset consists of those video that are correctly classified by the model with $N=2$ frames as input, but that were misclassified by the $N=1$ model, for $N=4$ we exclude the videos from $N=1$ and $N=2$, etc.

\end{document}